\documentclass[journal]{IEEEtran}
\ifCLASSINFOpdf
\else
\fi

\ifCLASSOPTIONcompsoc
  \usepackage[caption=false,font=normalsize,labelfont=sf,textfont=sf]{subfig}
\else
  \usepackage[caption=false,font=footnotesize]{subfig}
\fi

\usepackage{graphics} 
\usepackage{amsmath,amssymb,amsfonts}
\usepackage{pdfpages}
\usepackage{hyperref}
\usepackage[ruled,vlined]{algorithm2e}
\usepackage{booktabs}
\usepackage{multirow}
\usepackage{flushend}
\usepackage{comment}

\begin{document}

\title{Training Aware Sigmoidal Optimizer}

\author{David Macêdo*, \emph{Member, IEEE}, Pedro Dreyer*,\\Teresa Ludermir, \emph{Senior Member}, IEEE, and Cleber Zanchettin, \emph{Member, IEEE}
\thanks{$^{1}$*equal contributions. David Macêdo, Pedro Dreyer, Teresa Ludermir, and Cleber Zanchettin are with the Centro de Informática (CIn), Universidade Federal de Pernambuco (UFPE), Recife, Pernambuco, Brasil. Emails: {\tt\small \{dlm,phdl,cz,tbl\}@cin.ufpe.br}.}%
}

\maketitle

\begin{abstract}
Proper optimization of deep neural networks is an open research question since an optimal procedure to change the learning rate throughout training is still unknown. Manually defining a learning rate schedule involves troublesome time-consuming try and error procedures to determine hyperparameters such as learning rate decay epochs and learning rate decay rates. Although adaptive learning rate optimizers automatize this process, recent studies suggest they may produce overffiting and reduce performance when compared to fine-tuned learning rate schedules. Considering that deep neural networks loss functions present landscapes with much more saddle points than local minima, we proposed the Training Aware Sigmoidal Optimizer (TASO), which consists of a two-phases automated learning rate schedule. The first phase uses a high learning rate to fast traverse the numerous saddle point, while the second phase uses low learning rate to slowly approach the center of the local minimum previously found. We compared the proposed approach with commonly used adaptive learning rate schedules such as Adam, RMSProp, and Adagrad. Our experiments showed that TASO outperformed all competing methods in both optimal (i.e., performing hyperparameter validation) and suboptimal (i.e., using default hyperparameters) scenarios. 
\end{abstract}


\IEEEpeerreviewmaketitle

\section{INTRODUCTION}

\begin{figure}[!t]
\centering
\includegraphics[width=0.95\linewidth]{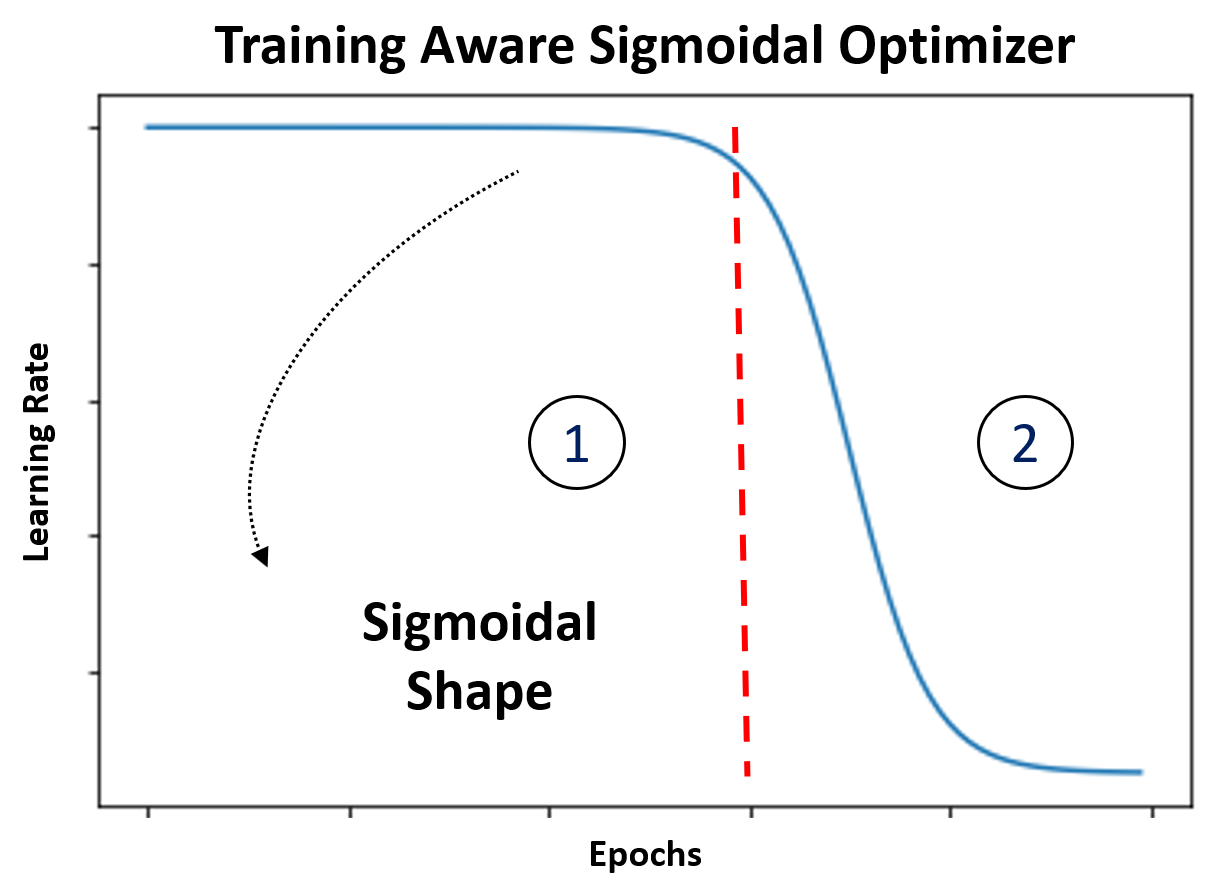}
\caption{\textmd{Inspired by the work of Dauphin et al. \cite{dauphin:saddle_point}, the TASO learning rate schedule comprises two distinct phases: During the first phase (1), the learning rate is kept constant for fast pass through the numerous saddle points. In the second phase (2), a rapid decrease in the learning rate allows for an efficient search around the local minimum found on the previous phase. We call TASO an \emph{automated} learning rate schedule because, other than default hyperparameters, only the initial learning rate needs to be provided to train the deep neural network. In other words, unlike usual learning rate schedules, it is \emph{not} necessary to \emph{manually} define \emph{well-tuned} learning rate decay epochs or learning rate decay rates. Therefore, using TASO is as straightforward as using \emph{adaptive learning rate} methods such as Adam, RMSProp, and Adagrad.}}
\label{fig:taso_general}
\end{figure}

\IEEEPARstart{D}{eep} neural networks are being used in many different fields with remarkable results.  Image classification and segmentation \cite{NIPS2019_9035, 9052469, xie2019self}, speech recognition \cite{saon:speech, park2020improved} and natural language processing (NLP) \cite{devlin:bert} are some fields where deep learning is producing state-of-the-art performance \cite{Alom2019}. However, while the results are encouraging, the ideal way to optimize such models are still unclear. The literature presents many optimizers with no clear dominating method across different tasks \cite{schmidt2020descending}.

One  essential hyperparameter common to all optimizers is the initial learning rate, which dictates how fast the model parameters are updated in the beginning of the training. The performance of the optimizer depends on its learning rate change strategy and finding the optimal way to achieve this is a fundamental research problem.

The change of the learning rate throughout training is usually obtained using \emph{learning rate schedule} or \emph{adaptive learning rate} optimizers \cite{10.5555/1953048.2021068}. On the one hand, in \emph{learning rate schedule} optimizers, the way the learning rate changes during training is defined before the training begins. On the other hand, \emph{adaptive learning rate} optimizers control the learning rate based on the training history, commonly using gradient information obtained during loss minimization.

Despite the increasing popularity and the relative success, adaptive learning rate optimizers (e.g., Adam \cite{kingma:adam}, RMSProp \cite{tieleman:rmsprop}, and Adagrad \cite{duchi:adagrad}), have shown marginal performance gains when compared to a \emph{manually well-tuned} learning rate schedule \cite{wilson:marginal}. Additionally, \cite{wilson:marginal} also showed that adaptive learning rate methods may produce overfitting and, consequently, reduced performance.

Indeed, \cite{wilson:marginal} showed that a \emph{manually well-tuned} learning rate schedules might perform similar to (or even better than) adaptive learning rate methods while avoiding their complexity and propensity to overfitting. However, constructing a manually well-tuned learning rate schedule is a troublesome procedure and demands many trials and errors, as it requires the definition of hyperparameters such as \emph{learning rate decay epochs} and \emph{learning rate decay rates}. Therefore, find enhanced optimizer is still an open and relevant research problem.

To combine the best of both approaches (the straightforwardness of the adaptive learning rate methods and the high performance of manually well-tuned learning rate schedules), we propose an \emph{automated} learning rate schedule method called Training Aware Sigmoidal Optimizer (TASO). Our approach is as straightforward as any adaptive learning rate method, as only the initial learning rate needs to be provided. Indeed, Similar to adaptive learning rate optimizers, we may use the TASO default hyperparameters, or we may perform validation to find optimal hyperparameters.

TASO is a two-phase learning rate schedule based mainly on the work of Dauphin et al. \cite{dauphin:saddle_point}. Based on the mentioned work, we speculate that the training of deep neural networks follows two phases. During the first phase, the optimizer needs to fast pass through a high amount of saddle points. After that, the optimizer finds a local minimum in which it needs to slowly converge to its center. A high learning rate would make traversing the saddle points plateaus faster in the initial stage of the training. Subsequently, once the optimization arrives near the local minimum vicinity, a fast decrease in the learning rate to lower values is more adequate to converge to this critical point slowly. We imposes this combined behavior by making the TASO learning rate schedule to follow a \emph{Sigmoidal shape} during the network training (Fig.~\ref{fig:taso_general}).

We compared TASO to the commonly used adaptive learning rate optimizers such as Adam \cite{kingma:adam}, RMSProp \cite{tieleman:rmsprop}, and Adagrad \cite{duchi:adagrad}. We used the mentioned optimizers to train the LeNet5 \cite{mnist}, VGG19 \cite{vgg}, and ResNet18 \cite{ResNet} models on the MNIST \cite{mnist}, CIFAR10 \cite{Krizhevsky09learningmultiple}, and CIFAR100 \cite{Krizhevsky09learningmultiple} datasets. Our experiments showed that TASO outperformed the compared adaptive learning rate optimizers in all combinations of models and datasets both for optimal (use of specific hyperparameters validate on the same dataset and model) and suboptimal (use of default hyperparameters validate on a different dataset and model) use cases.

\section{Background}

\begin{algorithm}[!t]

\textbf{Require:} Global learning rate $\epsilon$

\textbf{Require:} Initial parameters $\theta$

\textbf{Require:} Small constant $\delta$, normally $10^{-7}$, for numerical stability

\While{stopping criterion not met}
{
	Sample a minibatch of $n$ examples from the training set $\{x^{(1)},\dots,x^{(n)}\}$ with corresponding targets $y^{(i)}$.
	
	Compute gradient estimate: $g \leftarrow + \nabla_\theta \sum_i L(f(x^{(i)}; \theta), y^{(i)})$ 
	
	Accumulate squared gradient: $r \leftarrow r + g \odot g$
	
	Compute update: $ \Delta \theta \leftarrow - \frac{\epsilon}{\delta + \sqrt{r}} \odot g$
	
	Apply update: $\theta \leftarrow \theta + \Delta \theta$

}

\caption{Adagrad}
\label{alg:adagrad}
\end{algorithm}

\begin{algorithm}[!t]

\textbf{Require:} Global learning rate $\epsilon$, decay rate $\rho$

\textbf{Require:} Initial parameters $\theta$

\textbf{Require:} Small constant $\delta$, normally $10^{-6}$, for numerical stability

\While{stopping criterion not met}
{
	Sample a minibatch of $n$ examples from the training set $\{x^{(1)},\dots,x^{(n)}\}$ with corresponding targets $y^{(i)}$.
	
	Compute gradient estimate: $g \leftarrow + \nabla_\theta \sum_i L(f(x^{(i)}; \theta), y^{(i)})$ 
	
	Accumulate squared gradient: $r \leftarrow \rho r +(1-\rho ) g \odot g$
	
	Compute update: $ \Delta \theta \leftarrow - \frac{\epsilon}{ \sqrt{\delta r}} \odot g$
	
	Apply update: $\theta \leftarrow \theta + \Delta \theta$

}

\caption{RMSProp}
\label{alg:rmsprop}
\end{algorithm}

\begin{algorithm}[!t]

\textbf{Require:} Learning rate $\epsilon$

\textbf{Require:} Exponential decay rates for moment estimates, $\rho_1$ and $\rho_2$ in $[0,1)$

\textbf{Require:} Small constant $\delta$, normally $10^{-8}$, for numerical stability

\textbf{Require:} Initial parameters $\theta$ 

\While{stopping criterion not met}
{
	Sample a minibatch of $n$ examples from the training set $\{x^{(1)},\dots,x^{(n)}\}$ with corresponding targets $y^{(i)}$.
	
	Compute gradient estimate: $g \leftarrow + \nabla_\theta \sum_i L(f(x^{(i)}; \theta), y^{(i)})$ 
	
	$t \leftarrow t + 1$
	
	Update biased first moment estimate: $s \leftarrow \rho_1 s +(1-\rho_1 ) g$
	
	Update biased second moment estimate: $t \leftarrow \rho_2 r +(1-\rho_2 ) g \odot g$
	
	Correct bias in first moment: $\hat{s} \leftarrow \frac{s}{1-\rho_1^t}$
	
	Correct bias in second moment: $\hat{r} \leftarrow \frac{r}{1-\rho_2^t}$
	
	Compute update: $ \Delta \theta \leftarrow -\epsilon \frac{\hat{s}}{\sqrt{\hat{r}}+\delta}$
	
	Apply update: $\theta \leftarrow \theta + \Delta \theta$

}

\caption{Adam}
\label{alg:adam}
\end{algorithm}

Duchi et al. \cite{duchi:adagrad} proposed the Adagrad optimizer by scaling the global learning rate by the inverse square root of the sum of all squared values of the gradient. While having some theoretical properties for the convex optimization case, Adagrad does not perform so well for optimization of deep neural networks. The main issue seems to be the accumulative term, which is a monotonic increasing function that always increase. This can lead to an excessive decrease in the learning rate during later parts of the training. The Adagrad optimizer algorithm can be seen in the Algorithm \ref{alg:adagrad}.

Tieleman at al. \cite{tieleman:rmsprop} proposed the RMSProp optimizer as a modification of Adagrad by changing the accumulation of gradients into a weighted moving average, similar to the stochastic gradient descent (SGD) using momentum \cite{polyak:momentum}. The RMSProp optimizer algorithm can be seen in the Algorithm \ref{alg:rmsprop}. A variation of RMSProp was proposed in \cite{graves:rmsprop_centered}. In this version, the gradient is normalized by an estimation of its variance.

Kingma et al. \cite{kingma:adam} proposed the Adam optimizer by combining the RMSProp with the momentum used in SGD. The two methods are called first-moment terms and second-moment terms in the Adam optimizer. A bias correction term was included to account for the initialization of the momentum terms at zero. The Adam optimizer algorithm can be seen in the Algorithm \ref{alg:adam}. Recent research \cite{reddi:amsgrad} has found some theoretical shortcomings of the Adam optimizer, where the usage of exponential moving average caused non-convergence on a convex toy-problem. An alternative method, called AmsGrad, was then developed to overcome this deficiency.

\section{Training Aware Sigmoidal Optimizer}

\begin{figure}[!t]
	\centering
	\includegraphics[width=0.95\linewidth]{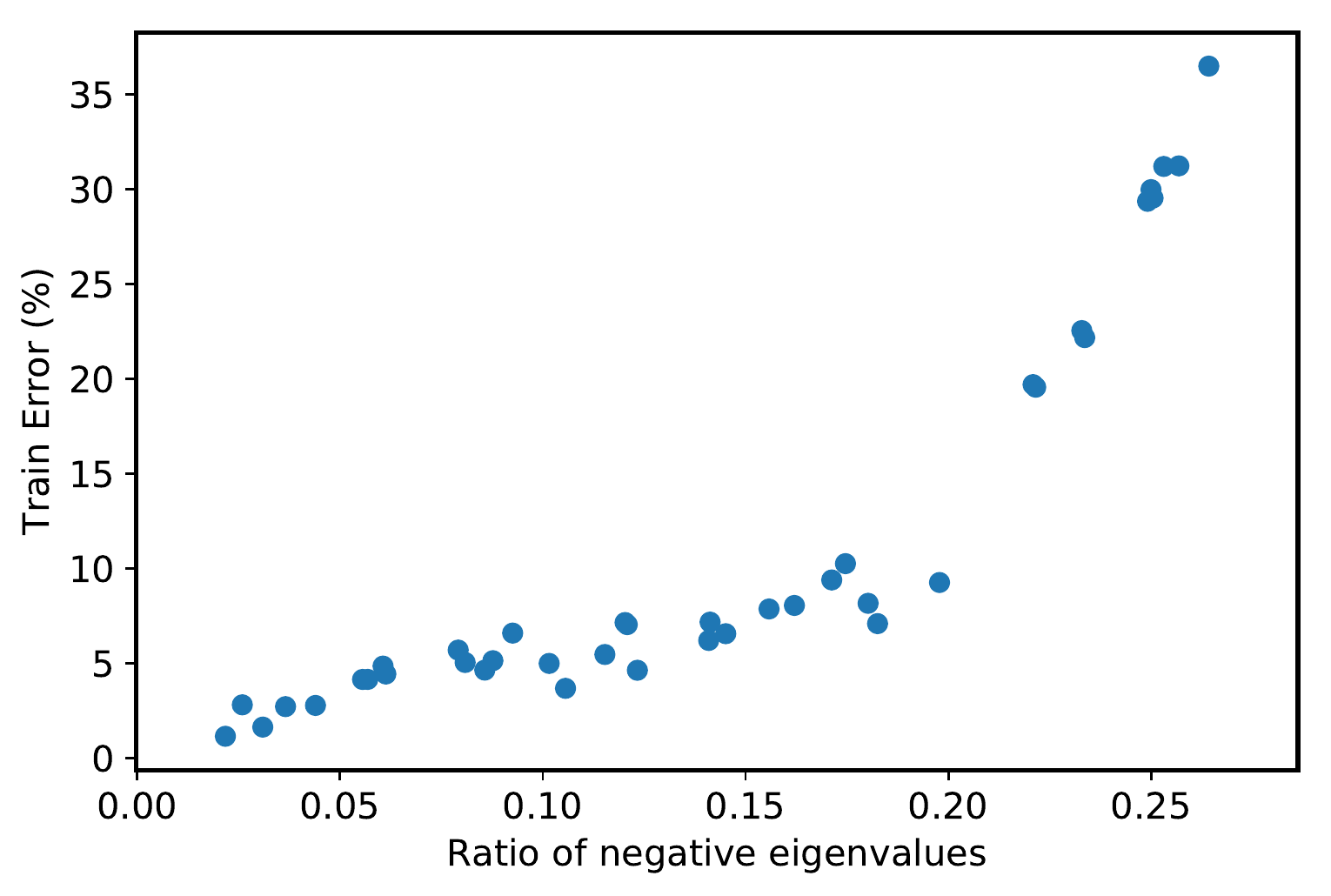}
	\caption{\textmd{Critical point training errors relative to its ratio of negative eigenvalues in its Hessian matrix. Adapted from  \cite{dauphin:saddle_point}}}
	\label{fig:critical_point_error}
\end{figure}

Training deep neural networks involves optimizing a high dimensional non-convex space. For decades, this fact was deemed a shortcoming of gradient descent, as it is local optimization approach. However, Dauphin et al. \cite{dauphin:saddle_point} showed that the training error of critical points (i.e., a maximum, minimum, or a saddle point) is correlated with the fraction of negative eigenvalues on their hessian matrices (Fig.~\ref{fig:critical_point_error}). Consequently, all local minima present similar training errors. Therefore, it does not matter much on which local minimum the optimizer got trapped. The mentioned work also demonstrated that the loss landscape of deep networks present much more saddle points than local minima.

Based on the previously mentioned observations, we speculate that the optimization of deep neural networks appear to happen in two distinct phases. Initially, the optimizer traverse the numerous saddle point and then it approaches a rare local minima. Tishby et al. \cite{7133169} appears to corroborate the idea that the training deep networks is composed of two different stages. They argue that during the ``fitting'' phase the mutual information between the hidden layers and the input increases and in the course of the ``compression'' phase the same mutual information decreases.

Considering the previously mentioned hypothesis, we argue that the learning rate should behave differently during these two training phases. In the initial portion of training, a high learning rate allows fast traversing the saddle points plateaus. Indeed, Dauphin et al. \cite{dauphin:saddle_point} showed that optimization near sadle points may be hard and therefore we propose to keep the initial high learning rate during this phase. Once the optimization arrives near a rare local minima vicinity, we propose to fast decrease the learning rate to allow fine-tuned local search in the neighborhood of the mentioned critical point.

To produce a high training rate during the first phase and a fast decreasing learning rates during the second phase, the main idea is to calculate the learning rate used in each epoch based on the \emph{Sigmoidal shape} curve Equation \eqref{eq:tas}.


\begin{equation}
\label{eq:tas}
\epsilon=\frac{\epsilon_i}{1+\exp\left(\alpha \left(\frac{k}{k_t}-\beta\right)\right)}+\epsilon_f 
\end{equation}

\begin{algorithm}[!t]

\textbf{Require:} initial learning rate $\epsilon_{i}$ and final learning rate $\epsilon_{f}$

\textbf{Require:} hyperparameters $\alpha$ and $\beta$

\textbf{Require:} Initial parameters $\theta$

\While{stopping criterion not met}
{
	Sample a minibatch of $n$ examples from the training set $\{x^{(1)},\dots,x^{(n)}\}$ with corresponding targets $y^{(i)}$.
	
	Compute gradient estimate: $g \leftarrow +\frac{1}{n} \nabla_\theta \sum_i L(f(x^{(i)}; \theta), y^{(i)})$ 
	
	Compute new learning rate: $\epsilon \leftarrow \frac{\epsilon_i}{1+\exp(\alpha (\frac{k}{k_t} - \beta))} + \epsilon_f$
	
	Apply update: $\theta \leftarrow \theta -\epsilon g$

}

\caption{TASO}
\label{alg:taso}
\end{algorithm}

In the previous equation, $\epsilon_i$ and $\epsilon_f$ are the initial and final learning rate, $k$ is the current epoch, and $k_t$ is the total number of epochs. Notice that, in order to know in which epoch the learning rate should start decreasing, the total epochs planed to be trained needs to be informed to the TASO equation. This is the reason we say our solution is \emph{training aware}. To the best of our knowledge, current learning rate schedule and adaptive learning rate optimizers do not make use of this easily available information to improve its overall performance. 

\begin{figure}[!t]
\centering
\includegraphics[width=0.95\linewidth]{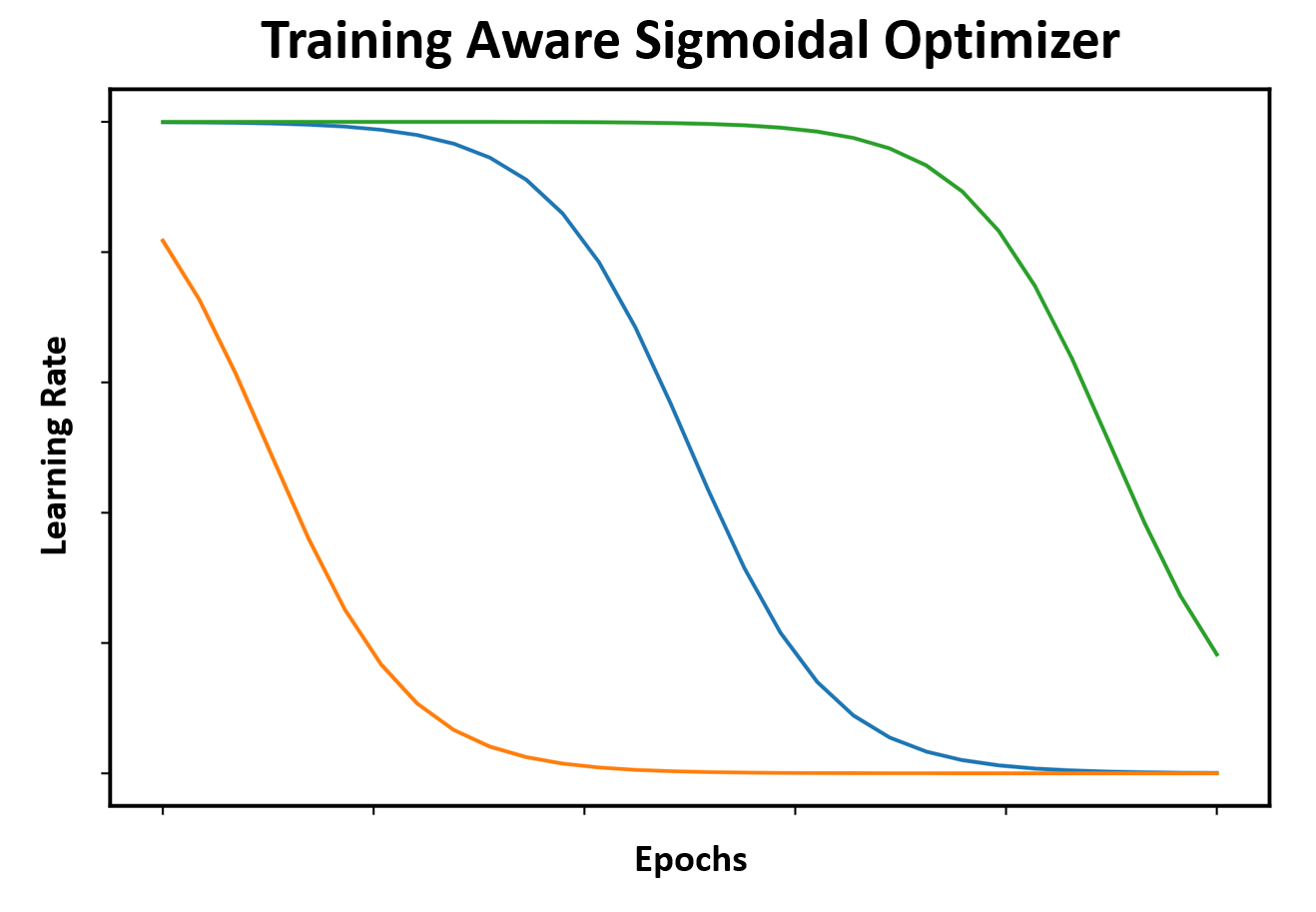}
\caption{\textmd{TASO curves as a function of hyperparameters $\alpha$ and $\beta$. The usual sigmoidal shape (blue line). $\alpha\beta<6$ (orange line) and $\alpha(1-\beta)<6$ (green line) represent combinations of $\alpha$ and $\beta$ values that should be avoided.}}
\label{fig:bad_taso}
\end{figure}

Additionally, the hyperparameter $\alpha$ control how fast the learning should decay after the first phase. Finally, the hyperparameter $\beta$ determines when the transition between the first and the second stage should occur. The TASO optimizer algorithm can be seen in the Algorithm \ref{alg:taso}. Fig.~\ref{fig:bad_taso} present the TASO curve shape for different combinations of hyperparameters. The Equations~(2-4) demonstrates how TASO produces essentially a constant initial learning rate $\epsilon_i$ during the first part of the training while the Equations~(5-7) shows how TASO approximates the much lower final learning rate through the second part of the optimization process.

\begin{align}
\label{eq:tas_early}
\epsilon_{(k=1)}&=\frac{\epsilon_i}{1+\exp\left(\alpha\left(\frac{1}{k_t}-\beta\right)\right)}+\epsilon_f\\ 
&\approx\frac{\epsilon_i}{1+\exp(-\alpha\beta)}+\epsilon_f\\
&\approx\epsilon_i
\end{align}

\begin{align}
\label{eq:tas_later}
\epsilon_{(k=k_f)}&=\frac{\epsilon_i}{1+\exp\left(\alpha\left(\frac{k_t}{k_t}-\beta\right)\right)}+\epsilon_f\\
&=\frac{\epsilon_i}{1+\exp(\alpha(1-\beta))}+\epsilon_f\\
&\approx\epsilon_f
\end{align}


The hyperparameters $\epsilon_i$ and $\epsilon_f$ do not present exactly the initial and final learning rates effectively used, as we have approximation in the previous equations. However, by restricting the choices of $\alpha$ and $\beta$, we can make these differences small enough to be negligible. As an useful heuristic, both $\alpha \beta$ and $\alpha (1-\beta)$ need to be higher than six to maintain errors below 5\%. Fig.~\ref{fig:bad_taso} shows examples of how the choice of non-conforming pairs of $\alpha$ and $\beta$ can create degenerative cases.

\section{Experiments}

We compared TASO with the more commonly used optimizers using different models and datasets. In the first set of experiments, we performed validation to found an optimal set of hyperparameters for a predefined pair of dataset and models (Subsection~\ref{optimal_case}). We call the hyperparameters found the \emph{default} hyperparameters for each optimizer. In the second set of experiments, we used the \emph{default} hyperparameters without validation to found suboptimal results for different pairs of datasets and models (Subsection~\ref{default_case}).

The study of suboptimal cases using default hyperparameters without performing validation on novel pair of datasets and models is relevant because it presents a measure of the robustness of the compared approaches. Additionally, it is relevant from a practical use perspective where we desire high quality results without necessarily performing a exhaustive hyperparameter search for each novel dataset and models used. Indeed, we usually prefer an optimizer that performs well using the default set hyperparameters. We repeated each experiment five times and reported the average and standard deviation. All reported accuracies and losses values were calculated on the test set. If not otherwise mentioned, we used 100 epochs for training. The source code is available\footnote{https://github.com/dlmacedo/training-aware-sigmoidal-optimizer}.

\subsection{Optimizers}

We used the same adaptive learning rate optimizers evaluated by Wilson et al. \cite{wilson:marginal}, which were Adagrad, RMSProp, and Adam. 
We also searched for the same set of initial learning rates of the mentioned work:\\


\begin{itemize}
\item \textbf{Adagrad}: $[0.1, 0.05, 0.01, 0.0075, 0.005]$
\item \textbf{RMSProp}: $[0.01, 0.005, 0.001, 0.0005, 0.0003, 0.0001]$
\item \textbf{Adam}: $[0.005, 0.001, 0.0005, 0.0003, 0.0001, 0.00005]$\\
\end{itemize}

\newpage

\begin{table}[!t]
\centering
\caption[caption]{Adagrad initial learning rate search for VGG19 on CIFAR10.}
\vskip -0.2cm
\begin{tabular*}{\linewidth}{@{\extracolsep{\fill}}lll@{}}
\toprule
 Learning Rate & Accuracy & Loss \\ 
\midrule
   0.1 & 89.18 ($\pm$0.16) & 0.45 ($\pm$0.01) \\ 
  0.05 & \textbf{89.52} ($\pm$0.31) & 0.45 ($\pm$0.01) \\ 
  0.01 & 88.77 ($\pm$0.09) & 0.46 ($\pm$0.01) \\ 
  0.0075 & 88.09 ($\pm$0.02) & 0.47 ($\pm$0.01) \\ 
  0.005 & 87.80 ($\pm$0.09) & 0.45 ($\pm$0.01) \\ 
\bottomrule 
\end{tabular*}
\label{tab:hyper_adagrd}
\centering
\vskip 0.2cm
\caption[caption]{RMSProp initial learning rate and variant search\\for VGG19 on CIFAR10.}
\vskip -0.2cm
\begin{tabular*}{\linewidth}{@{\extracolsep{\fill}}llll@{}}
\toprule
Optimizer & Learning Rate & Accuracy & Loss \\ 
\midrule
\multirow{6}{*}{RMSProp}
 & 0.01 & 89.94 ($\pm$0.13) & 0.44 ($\pm$0.02) \\ 
 & 0.005 & 90.10 ($\pm$0.32) & 0.42 ($\pm$0.01) \\ 
 & 0.001 & 90.74 ($\pm$0.01) & 0.40 ($\pm$0.01) \\ 
 & 0.0005 & \textbf{90.77} ($\pm$0.07) & 0.39 ($\pm$0.02) \\ 
 & 0.0003 & 90.54 ($\pm$0.15) & 0.40 ($\pm$0.01) \\ 
 & 0.0001 & 88.75 ($\pm$0.39) & 0.46 ($\pm$0.01) \\ 
\midrule 
\multirow{4}{*}{RMSProp centered}
 & 0.01 & 89.96 ($\pm$0.17) & 0.43 ($\pm$0.01) \\ 
 & 0.005 & 90.00 ($\pm$0.17) & 0.42 ($\pm$0.01) \\ 
 & 0.001 & 90.65 ($\pm$0.26) & 0.39 ($\pm$0.03) \\ 
 & 0.0005 & \textbf{90.76} ($\pm$0.21) & 0.38 ($\pm$0.02) \\ 
 & 0.0003 & 90.58 ($\pm$0.16) & 0.38 ($\pm$0.03) \\ 
 & 0.0001 & 88.72 ($\pm$0.38) & 0.46 ($\pm$0.01) \\ 
\bottomrule 
\end{tabular*}
\label{tab:hyper_rmsprop}
\centering
\vskip 0.2cm
\caption[caption]{Adam initial learning rate and variant search\\for VGG19 on CIFAR10.}
\vskip -0.2cm
\begin{tabular*}{\linewidth}{@{\extracolsep{\fill}}llll@{}}
\toprule
Optimizer & Learning Rate & Accuracy & Loss \\ 
\midrule
\multirow{6}{*}{Adam}
 & 0.005 & 90.17 ($\pm$0.05) & 0.39 ($\pm$0.01) \\ 
 & 0.001 & 90.93 ($\pm$0.34) & 0.38 ($\pm$0.02) \\ 
 & 0.0005 & \textbf{91.02} ($\pm$0.04) & 0.37 ($\pm$0.01) \\ 
 & 0.0003 & 90.77 ($\pm$0.18) & 0.37 ($\pm$0.01) \\ 
 & 0.0001 & 88.96 ($\pm$0.09) & 0.44 ($\pm$0.01) \\ 
 & 0.00005 & 86.54 ($\pm$0.23) & 0.52 ($\pm$0.02) \\ 
\midrule 
\multirow{6}{*}{Adam AmsGrad}
 & 0.005 & 90.38 ($\pm$0.14) & 0.40 ($\pm$0.01) \\ 
 & 0.001 & 91.01 ($\pm$0.34) & 0.37 ($\pm$0.01) \\ 
 & 0.0005 & \textbf{91.33} ($\pm$0.15) & 0.36 ($\pm$0.01) \\ 
 & 0.0003 & 91.06 ($\pm$0.27) & 0.37 ($\pm$0.02) \\ 
 & 0.0001 & 88.66 ($\pm$0.11) & 0.45 ($\pm$0.01) \\ 
 & 0.00005 & 86.40 ($\pm$0.09) & 0.53 ($\pm$0.01) \\ 
\bottomrule 
\end{tabular*}
\label{tab:hyper_adam}

\end{table}
\begin{table}[!t]
\centering
\caption[caption]{TASO initial learning rate search\\for VGG19 on CIFAR10. Moment equal to 0.9.}
\vskip -0.2cm
\begin{tabular*}{\linewidth}{@{\extracolsep{\fill}}llll@{}}
\toprule
Optimizer & Learning Rate* & Accuracy & Loss \\ 
\midrule
\multirow{7}{*}{SGD}
 & 2 & 35.91 ($\pm$36.64) & 1.69 ($\pm$0.87) \\ 
 & 1 & 88.94 ($\pm$0.18) & 0.47 ($\pm$0.01) \\ 
 & 0.5 & 89.72 ($\pm$0.42) & 0.42 ($\pm$0.03) \\ 
 & 0.25 & \textbf{90.39} ($\pm$0.29) & 0.41 ($\pm$0.01) \\ 
 & 0.05 & 89.66 ($\pm$0.18) & 0.44 ($\pm$0.02) \\ 
 & 0.01 & 86.00 ($\pm$0.14) & 0.52 ($\pm$0.01) \\ 
 & 0.001 & 78.78 ($\pm$0.10) & 0.62 ($\pm$0.01) \\ 
\midrule 
\multirow{7}{*}{SGD with Momentum}
 & 2 & 10.05 ($\pm$0.03) & 2.48 ($\pm$0.15) \\ 
 & 1 & 10.04 ($\pm$0.03) & 2.33 ($\pm$0.01) \\ 
 & 0.5 & 25.82 ($\pm$22.38) & 1.94 ($\pm$0.53) \\ 
 & 0.25 & 55.32 ($\pm$33.12) & 1.33 ($\pm$0.74) \\ 
 & 0.05 & \textbf{91.01} ($\pm$0.13) & 0.37 ($\pm$0.00) \\ 
 & 0.01 & 90.55 ($\pm$0.11) & 0.40 ($\pm$0.01) \\ 
 & 0.001 & 86.38 ($\pm$0.13) & 0.50 ($\pm$0.01) \\ 
\midrule 
\multirow{7}{*}{SGD with Nesterov}
 & 2 & 10.14 ($\pm$0.10) & 2.32 ($\pm$0.00) \\ 
 & 1 & 10.08 ($\pm$0.09) & 2.31 ($\pm$0.00) \\ 
 & 0.5 & 10.02 ($\pm$0.02) & 2.30 ($\pm$0.00) \\ 
 & 0.25 & 28.70 ($\pm$26.22) & 1.93 ($\pm$0.54) \\ 
 & 0.05 & 90.00 ($\pm$0.71) & 0.41 ($\pm$0.01) \\ 
 & 0.01 & \textbf{90.49} ($\pm$0.28) & 0.40 ($\pm$0.01) \\ 
 & 0.001 & 86.56 ($\pm$0.22) & 0.51 ($\pm$0.00) \\ 
\bottomrule 
\end{tabular*}
\begin{flushleft}
\vskip -0.1cm
*The learning rate was kept constant during training.
\end{flushleft}
\label{tab:hyper_taso_1}
\centering
\vskip 0.2cm
\caption[caption]{TASO variant search for VGG19 on CIFAR10.}
\vskip -0.2cm
\begin{tabular*}{\linewidth}{@{\extracolsep{\fill}}llll@{}}
\toprule
 $\alpha$ & $\beta$ & Acurracy & Loss \\ 
\midrule
 10 & 0.3 & 90.96 ($\pm$0.12) & 0.37 ($\pm$0.01) \\ 
 10 & 0.5 & 91.66 ($\pm$0.31) & 0.38 ($\pm$0.01) \\ 
 10 & 0.7 & 91.97 ($\pm$0.19) & 0.38 ($\pm$0.02) \\ 
 25 & 0.3 & 90.73 ($\pm$0.17) & 0.37 ($\pm$0.01) \\ 
 25 & 0.5 & 91.61 ($\pm$0.27) & 0.37 ($\pm$0.01) \\ 
 25 & 0.7 & \textbf{91.98} ($\pm$0.19) & 0.35 ($\pm$0.01) \\ 
 50 & 0.3 & 90.85 ($\pm$0.30) & 0.36 ($\pm$0.01) \\ 
 50 & 0.5 & 91.94 ($\pm$0.04) & 0.37 ($\pm$0.02) \\ 
 50 & 0.7 & 91.95 ($\pm$0.25) & 0.37 ($\pm$0.01) \\ 
\bottomrule 
\end{tabular*}
\label{tab:hyper_taso_2}
\end{table}

\begin{table}[!t]
\centering
\caption[caption]{Final results using optimal (validated) hyperparameters.\\VGG19 model on CIFAR10 dataset.}
\vskip -0.2cm
\begin{tabular*}{\linewidth}{@{\extracolsep{\fill}}lll@{}}
\toprule
 Optimizer & Test Acurracy & Loss \\ 
\midrule
Adagrad & 89.40 ($\pm$0.31) & 0.42 ($\pm$0.01) \\ 
RMSProp & 90.77 ($\pm$0.01) & 0.41 ($\pm$0.01) \\ 
Adam & 91.33 ($\pm$0.16) & 0.37 ($\pm$0.01) \\ 
TASO & \textbf{91.98} ($\pm$0.19) & 0.36 ($\pm$0.02) \\ 
\bottomrule 
\end{tabular*}
\label{tab:cifar10_vgg_best}
\centering
\vskip 0.2cm
\caption[caption]{Final results using optimal (validated) hyperparameters.\\VGG19 model on CIFAR10 dataset.\\Training during 25 epochs.}
\vskip -0.2cm
\begin{tabular*}{\linewidth}{@{\extracolsep{\fill}}lll@{}}
\toprule
Optimizer & Test Acurracy & Loss \\ 
\midrule
Adagrad & 86.10 ($\pm$0.27) & 0.44 ($\pm$0.01) \\ 
RMSProp & 88.20 ($\pm$0.33) & 0.41 ($\pm$0.01) \\ 
Adam & 88.56 ($\pm$0.16) & 0.38 ($\pm$0.00) \\ 
TASO & \textbf{90.02} ($\pm$0.41) & 0.34 ($\pm$0.01) \\ 
\bottomrule 
\end{tabular*}
\label{tab:cifar10_vgg_best_25epochs}
\centering
\vskip 0.2cm
\caption[caption]{Final results using suboptimal (default) hyperparameters.\\VGG19 model on CIFAR100 dataset.}
\vskip -0.2cm
\begin{tabular*}{\linewidth}{@{\extracolsep{\fill}}lll@{}}
\toprule
Optimizer & Test Acurracy & Loss \\ 
\midrule
Adagrad & 1.33 ($\pm$0.06) & 4.77 ($\pm$0.17) \\ 
RMSProp & 55.02 ($\pm$0.54) & 2.04 ($\pm$0.05) \\ 
Adam & 61.46 ($\pm$0.13) & 1.90 ($\pm$0.04) \\ 
TASO & \textbf{65.08} ($\pm$0.47) & 1.76 ($\pm$0.01) \\ 
\bottomrule 
\end{tabular*}
\label{tab:cifar100_vgg_best}
\centering
\vskip 0.2cm
\caption[caption]{Final results using suboptimal (default) hyperparameters.\\Resnet18 model on CIFAR10 dataset.}
\vskip -0.2cm
\begin{tabular*}{\linewidth}{@{\extracolsep{\fill}}lll@{}}
\toprule
Optimizer & Test Acurracy & Loss \\ 
\midrule
Adagrad & 19.03 ($\pm$0.78) & 2.19 ($\pm$0.01) \\ 
RMSProp & 92.13 ($\pm$0.28) & 0.37 ($\pm$0.01) \\ 
Adam & 92.35 ($\pm$0.29) & 0.35 ($\pm$0.01) \\ 
TASO & \textbf{93.15} ($\pm$0.08) & 0.35 ($\pm$0.01) \\ 
\bottomrule 
\end{tabular*}
\label{tab:cifar10_resnet_best}
\centering
\vskip 0.2cm
\caption[caption]{Final results using suboptimal (default) hyperparameters.\\Lenet5 model on MNIST dataset.}
\vskip -0.2cm
\begin{tabular*}{\linewidth}{@{\extracolsep{\fill}}lll@{}}
\toprule
Optimizer & Test Acurracy & Loss \\
\midrule
Adagrad & 75.16 ($\pm$8.01) & 1.15 ($\pm$0.36) \\ 
RMSProp & \textbf{99.09} ($\pm$0.03) & 0.03 ($\pm$0.01) \\ 
Adam & 99.03 ($\pm$0.03) & 0.03 ($\pm$0.01) \\ 
TASO & \textbf{99.09} ($\pm$0.03) & 0.03 ($\pm$0.01) \\ 
\bottomrule 
\end{tabular*}
\label{tab:mnist_lenet_best}
\end{table}

For the RMSProp optimizer, we opted to use the default value of $0.99$ for the exponential decay rate, as it seems to have little influence on the overall result. We tried the alternate version proposed by Graves at al. \cite{graves:rmsprop_centered}, which we called RMSProp centered. Besides the original Adam, we evaluate its alternative version, AmsGrad. For both Adam and AmsGrad, we left the exponential decay terms $\rho_1$ and $\rho_2$ at their default values of 0.9 and 0.99 as, similar to the RMSProp, changes in those values do not seem to impact training meaningfully.

For TASO, we investigated the initial learning rates 2, 1, 0.5, 0.25, 0.05, 0.01, and 0.001. We study the training without moment, with moment equals to 0.9, and with Nesterov moment equals to 0.9. Subsequently, we searched for the variants \mbox{$\alpha\!=\![10, 25, 50]$} and $\beta\!=\![0.3, 0.5, 0.7]$, which seems to encompass a satisfactory range of possible configurations. As we initially did not know optimal $\alpha$ and $\beta$, we validated those values using the correspond initial learning rate constant throughout training.

Naturally, in future works, the search for the best initial learning rate for TASO should use \emph{default} values for $\alpha$ and $\beta$ (which we subsequently found to be 25 and 0.7, respectively) rather than keep the candidate initial learning rates constants throughout training. We recommend using the final learning rate 20 times smaller than the initial learning rate.

\subsection{Databases}

The experiments were performed using MNIST \cite{mnist}, CIFAR10 \cite{Krizhevsky2009LearningImages}, and CIFAR100 \cite{Krizhevsky2009LearningImages}. All of them are image classification datasets with different degrees of complexity and size. MNIST is a dataset of handwriting digits composed of 70,000 greyscale images. The training set has 60,000 examples and a test set is composed of 10,000 images. The images have a size of 28 pixels by 28 pixels, with the digits being normalized and centralized. CIFAR10 has ten classes, and CIFAR100 has 100 categories. They both have 60,000 color images divided between 50,000 training images and 10,000 test images. The images have a size of 32 pixels by 32 pixels.

\subsection{Models}

We used three models in the experiments. LeNet5 \cite{mnist}, VGG19 \cite{vgg}, and ResNet18 \cite{ResNet}. The LeNet was developed in 1998 to identify handwriting digits. The VGG was created in 2014 and was one of the runner ups of the ILSVRC 2014 competition. ResNet was created in 2015 and won several image tasks competitions such as the ILSVRC 2015 image detection and localization.

\section{Results and Discussion}

This section is divided into two subsections. In the first subsection, we perform a grid search validation to find optimal hyperparameters for VGG19 on CIFAR10 using all optimization methods. In the second subsection, we investigate how well the \emph{previously found (default)} hyperparameters generalize to other datasets and models avoiding validation. Therefore, in this second subsection, we study the robustness of the competing methods using \emph{default (global)} hyperparameters.

\begin{figure*}[!t]
\centering
\subfloat[]{\includegraphics[width=0.33\linewidth]{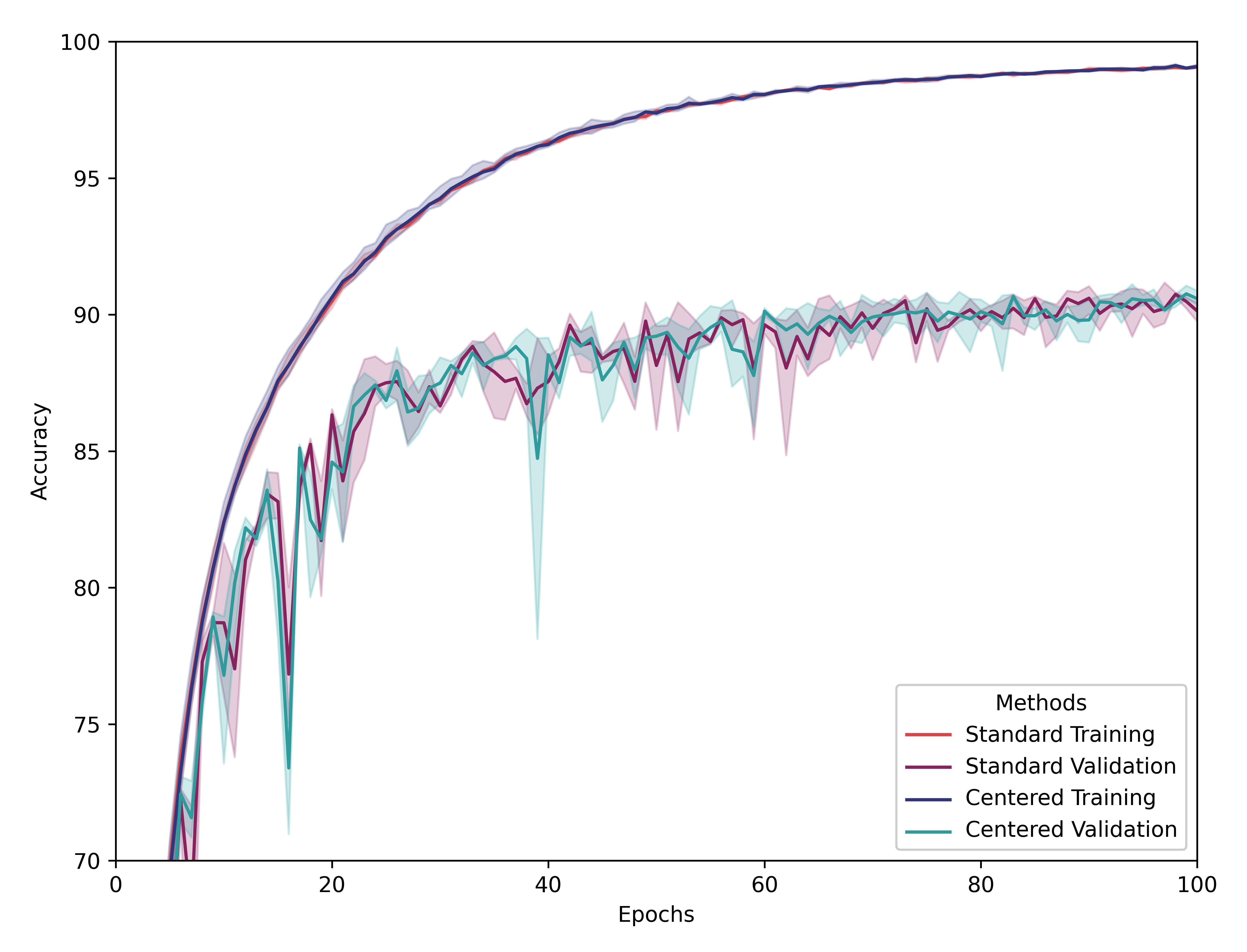}}
\subfloat[]{\includegraphics[width=0.33\linewidth]{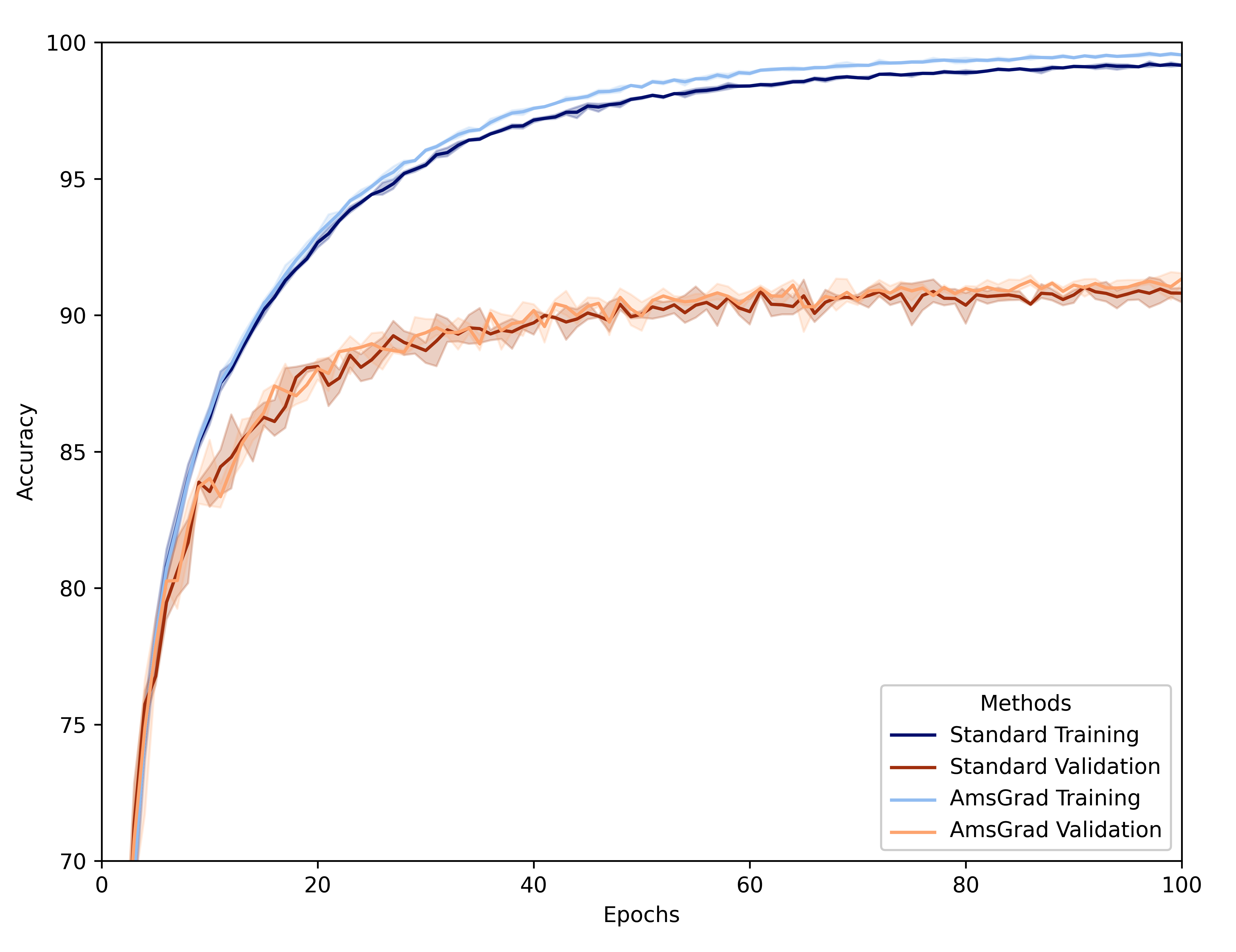}}
\subfloat[]{\includegraphics[width=0.33\linewidth]{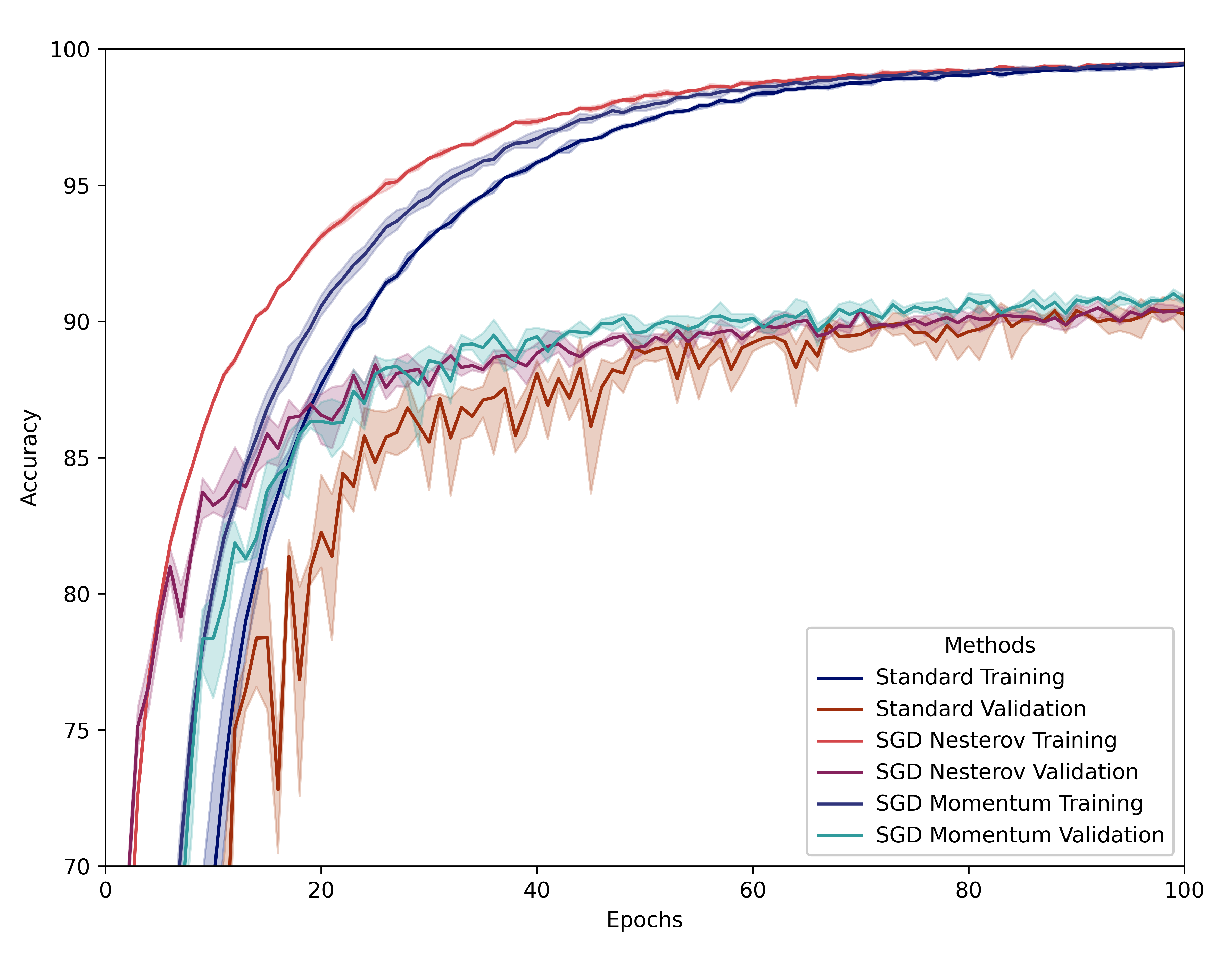}}
\caption[caption]{Validations of hyperparameters for VGG19 on CIFAR10. Solid lines represent the mean behavior of many runs of the same experiment. Shadow areas represent their variation. (a) RMSProp: non-centered worked best. (b) Adam: AmsGrad variant worked best. (c) TASO: non-Nesterov worked best.}
\label{fig:hyper_all}
\end{figure*}

\begin{figure*}[!t]
\centering
\subfloat[]{\includegraphics[width=0.475\linewidth]{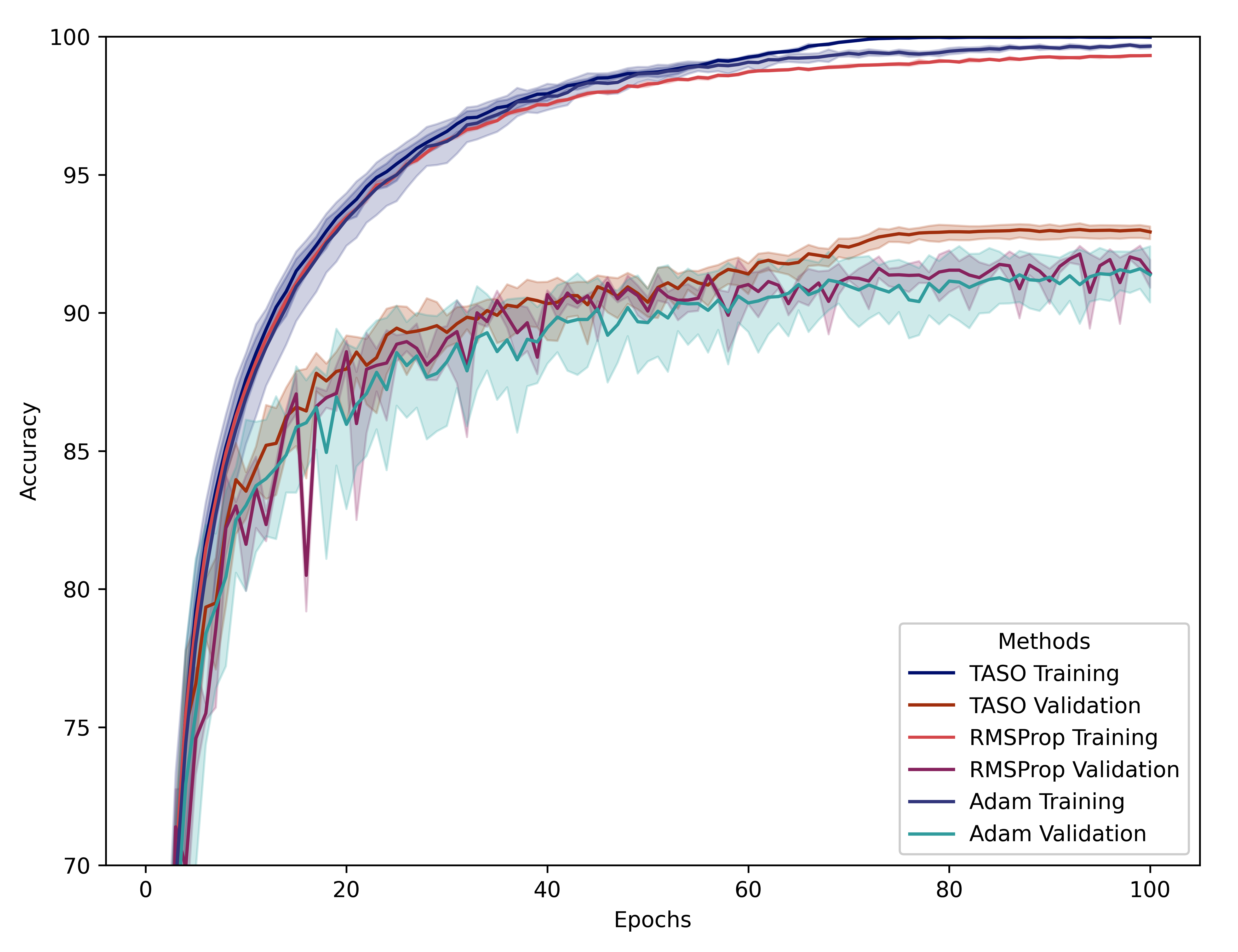}}
\subfloat[]{\includegraphics[width=0.475\linewidth]{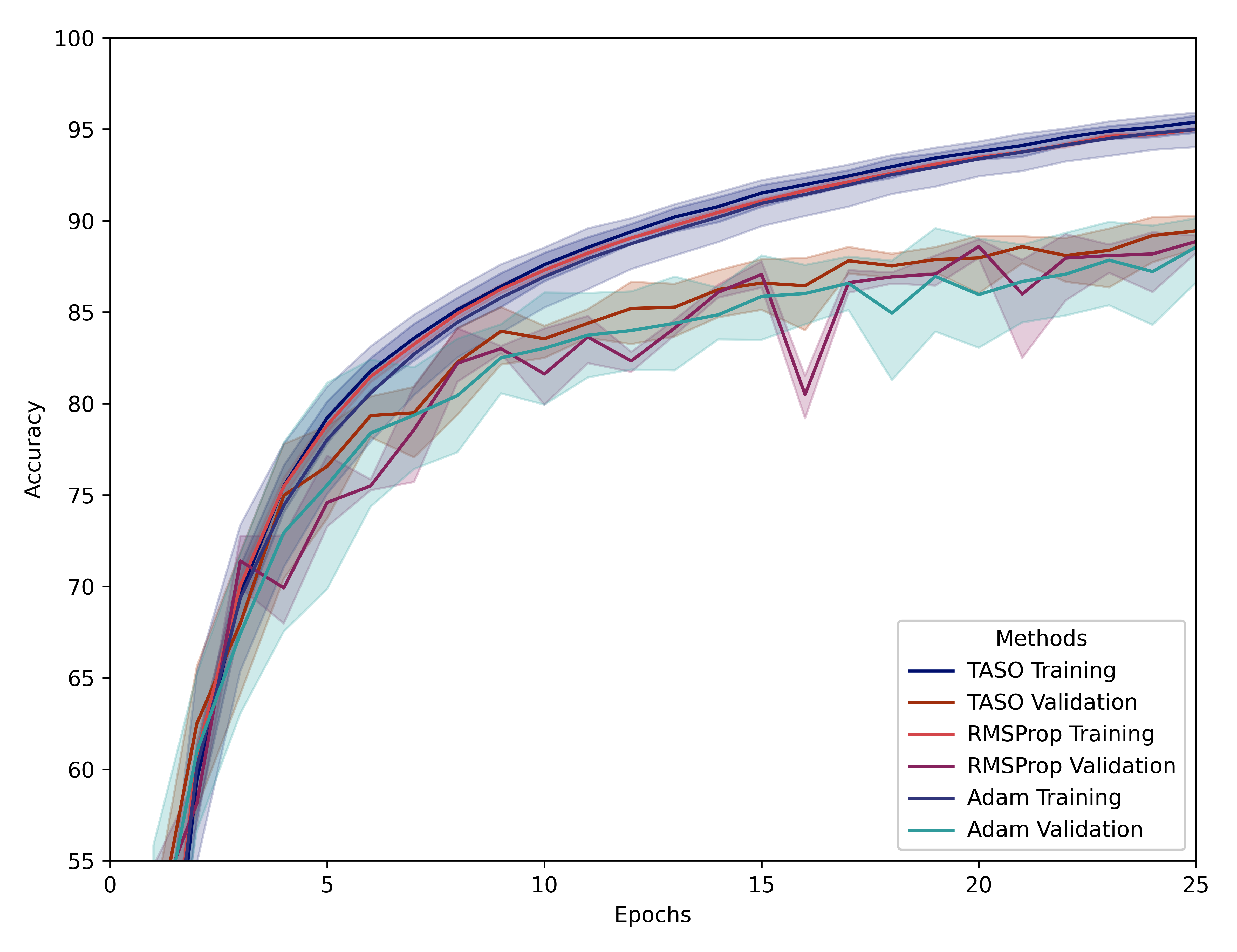}}
\caption{Final Results for VGG19 on CIFAR10 using optimal (validated) hyperparameters. Solid lines represent the mean behavior of many runs of the same experiment. Shadow areas represent their variation. (a) 100 epochs. (b) 25 epochs.}
\label{fig:cifar10_resnet_all_best}
\end{figure*}

\subsection{Comparison using optimal (validated) hyperparameters}\label{optimal_case}

For the Adagrad method, an initial learning rate of 0.05 performed best and therefore was used in subsequent experiments (Table~\ref{tab:hyper_adagrd}). For the RMSProp optimizer, the Table~\ref{tab:hyper_rmsprop} presents the best results for each hyperparameter set. Fig.~\ref{fig:hyper_all}a) compares the best results for centered and non-centered versions. The learning rates from 0.0003 to 0.001 presented similar results across both optimizers, and there is little difference between the centered and non-centered version. The non-centered version, with a learning rate of 0.00005, was used for further tests since it presented the best results. 

For the Adam optimizer, a summary of the results is shown in Table~\ref{tab:hyper_adam}. Fig.~\ref{fig:hyper_all}b) shows the best result of each version. From the experiments, we noted that the learning rates 0.001, 0.0005, and 0.0003 present very close results. While the AmsGrad version showed a better training accuracy, its test accuracy appears to be very similar to the original Adam optimizer. For further tests, the Adam AmsGrad optimizer was used with a default initial learning rate of 0.0005.

For the TASO optimizer, initially we concluded that the optimal initial learning rate was 0.05 using non-Nesterov moment of 0.9 (Table~\ref{tab:hyper_taso_1}). See also Fig.~\ref{fig:hyper_all}c). Subsequently, we searched for optimal $\alpha$ and $\beta$. The training results across the hyperparameters configurations can be seen in Table~\ref{tab:hyper_taso_2}. According to the results, we see a small difference varying $\alpha$ and $\beta$. The results show that TASO is robust for a wide range of values for $\alpha$ and $\beta$.

The best performance was achieved using $\alpha$ equals to 25 and $\beta$ equal to 0.7, which were defined as the default hyperparameters for TASO. It was also verified that the new calculations added in the TASO optimizer did not interfere with the training time, as the backpropagation is much more computationally intensive than calculating the new learning rate each epoch.

After finding the default hyperparameters for all optimizer, we re executed all experiments to obtain the final results for VGG19 on CIFAR10 presnted in Fig.~\ref{fig:cifar10_resnet_all_best}a) and Table~\ref{tab:cifar10_vgg_best}. There are optimal results, as the default hyperparameters were obtained using the same model and dataset. Experiments were also performed using only 25 epochs. The result can be found in Table~\ref{tab:cifar10_vgg_best_25epochs} and Fig.~\ref{fig:cifar10_resnet_all_best}b). The TASO method achieved the best results in both cases. In both experiments, we can easily visualize a bump in the accuracy close to the epochs 20 (25 epochs case) and 70 (100 epochs case), which correspond to the moment the learning rate started to decrease fast.

\subsection{Comparison using suboptimal (default) hyperparameters}\label{default_case}

\subsubsection{VGG19 and CIFAR100}

\begin{figure}[!t]
\centering
\subfloat[]{\includegraphics[width=0.95\linewidth]{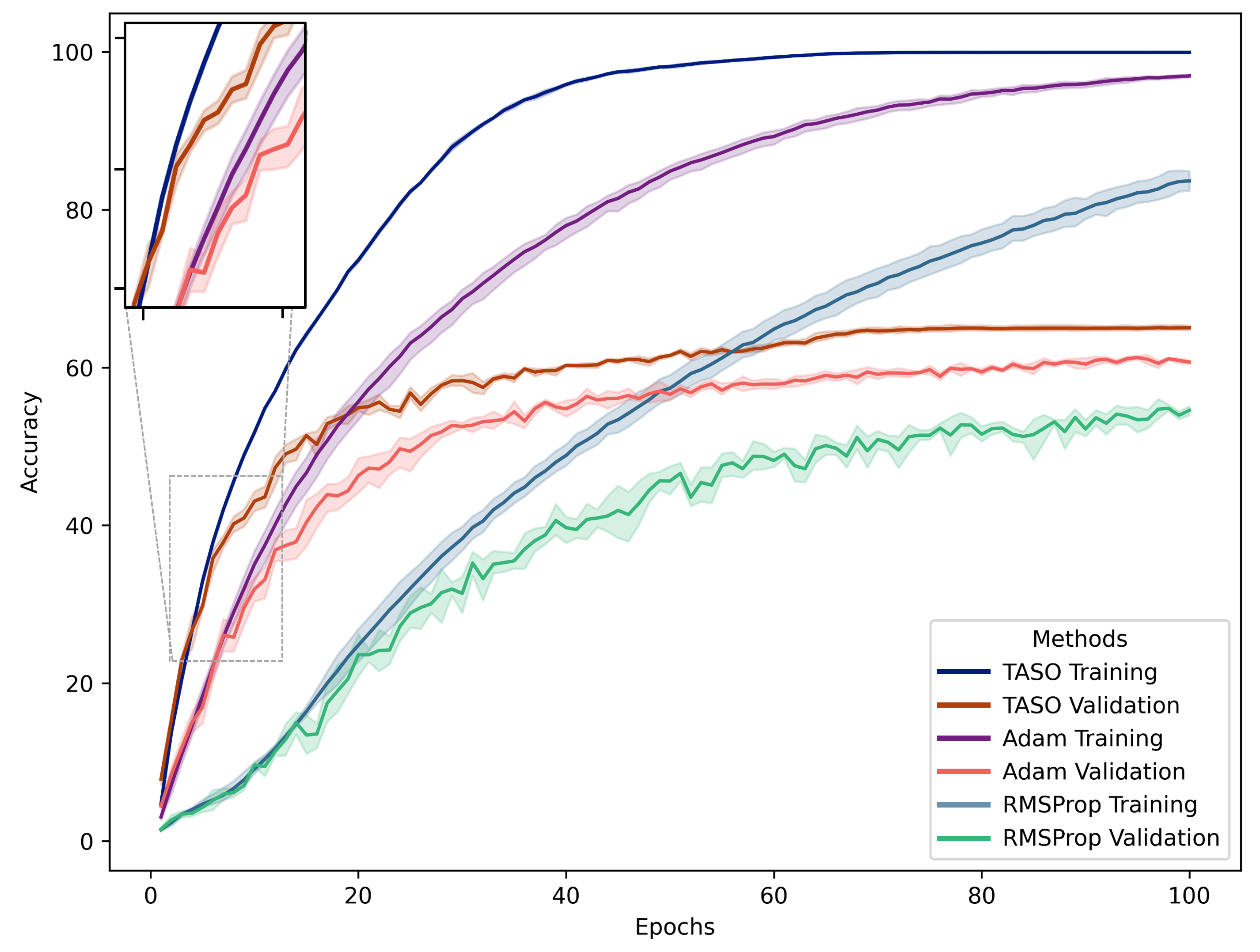}}
\caption{Final results for VGG19 on CIFAR100 using suboptimal (default) hyperparameters. Solid lines represent the mean behavior of many runs of the same experiment. Shadow areas represent their variation.}
\label{fig:cifar100_vgg_all_best}
\end{figure}

Using the best hyperparameters from the previous experiment, the same model was evaluated on the CIFAR100 dataset. The results can be seen in Table \ref{tab:cifar100_vgg_best} and Fig.~\ref{fig:cifar100_vgg_all_best}. We note once again that the TASO has the best overall result. Note as well that the Adagrad failed to converge, indicating the non-reliability of the optimizer.

\subsubsection{ResNet18 on CIFAR10}

The results of changing the model but maintained the CIFAR10 datasets can be seen in Table \ref{tab:cifar10_resnet_best}. 
Those are so far the best results, where the TASO method archives more than 2\% from the second-best performing optimizer. This more significant improvement could come from the fact that ResNet has a more complex deep model than VGG19, which could have a more complex loss function space, making it harder to optimize without changing the learning rate. Note that Adagrad, while it did converge this time, this optimizer presented much lower accuracy than the other methods.

\subsubsection{MNIST and LeNet5}

The last test, which compares the LeNet5 model trained on the MNIST dataset, is shown in Table \ref{tab:mnist_lenet_best}. The results demostrated that all the methods except for Adagrad managed to archive near-perfect accuracy and thus could be considered equivalent to the task of best training this particular dataset and model.

\section{Conclusion}

In this paper, we proposed TASO, a novel deep neural networks optimizer. Based on recent theoretical advances in the understanding of deep neural networks loss functions landscape, TASO presents two phase. In the first training stage the initial learning rate is essentially kept constant while in the second training stage the learning rate drops abruptly. The transition from the first to the second phase is allowed by know in advance the total number of epochs planned to be trained.  

TASO is what we call an \emph{automated} learning rate schedule. Therefore, similarly to adaptive learning rate optimizer, it may be used in two ways. In the optimal case, hyperparameters validation is performed to achieve the best possible performance. In the suboptimal scenario, default hyperparameters are used. Differently from \emph{manually well-tuned} learning rate schedules, TASO does no require to find hyperparameters such as \emph{learning rate decay epochs} or the \emph{learning rate decay rates}.

Out experiments showed that TASO outperformed all adaptive learning rate optimizer in both optimal and suboptimal scenarios. We believe this fact may be understood as further evidence that adaptive learning rate optimizers may present propensity for overfitting. In future works, we plan to use TASO in other computer vision tasks (e.g., objection detection and semantic segmentation) and diferents research areas such as natural language processing and speech recognition.


\section*{Acknowledgment}

This work was supported in part by CNPq and FACEPE (Brazilian research agencies). We gratefully acknowledge the support of NVIDIA Corporation with the donation of the Titan XP GPU used for this research. 


\bibliographystyle{ieeetr}
\bibliography{references.bib}

\end{document}